\documentclass[letterpaper]{article} % DO NOT CHANGE THIS
\usepackage{aaai25}  % DO NOT CHANGE THIS
\usepackage{times}  % DO NOT CHANGE THIS
\usepackage{helvet}  % DO NOT CHANGE THIS
\usepackage{courier}  % DO NOT CHANGE THIS
\usepackage[hyphens]{url}  % DO NOT CHANGE THIS
\usepackage{graphicx} % DO NOT CHANGE THIS
\usepackage{natbib}  % DO NOT CHANGE THIS AND DO NOT ADD ANY OPTIONS TO IT
\usepackage{caption} % DO NOT CHANGE THIS AND DO NOT ADD ANY OPTIONS TO IT
\usepackage{algorithm}
\usepackage{algorithmic}

\usepackage{multirow}
\usepackage[table]{xcolor}
\usepackage{amssymb}
\usepackage{utfsym}
\usepackage{booktabs}
\usepackage{newfloat}
\usepackage{listings}
\DeclareCaptionStyle{ruled}{labelfont=normalfont,labelsep=colon,strut=off} % DO NOT CHANGE THIS
\floatstyle{ruled}
\newfloat{listing}{tb}{lst}{}
\floatname{listing}{Listing}
\title{Taylor Series-Inspired Local Structure Fitting Network for Few-shot Point Cloud Semantic Segmentation}
\author{
    Changshuo Wang\textsuperscript{\rm 1}, 
    Shuting He\textsuperscript{\rm 2}, 
    Xiang Fang\textsuperscript{\rm 3}$^*$, 
    Meiqing Wu\textsuperscript{\rm 1}$^*$, 
    Siew-Kei Lam\textsuperscript{\rm 1,3}\thanks{Corresponding Authors.}, 
    Prayag Tiwari\textsuperscript{\rm 4}
}
\affiliations{
    \textsuperscript{\rm 1}Cyber Security Research Center (CYSREN), Nanyang Technological University, Singapore\\
    \textsuperscript{\rm 2}Shanghai University of Finance and Economics, China\\

    \textsuperscript{\rm 3}College of Computing and Data Science, Nanyang Technological University, Singapore\\
    \textsuperscript{\rm 4}School of Information Technology, Halmstad University, Sweden \\
    wangchangshuo1@gmail.com, shuting.he@sufe.edu.cn, xfang9508@gmail.com, meiqingwu@ntu.eud.sg, assklam@ntu.edu.sg, prayag.tiwari@ieee.org
}

\usepackage{bibentry}
\begin{document}

\maketitle

\begin{abstract}
	Few-shot point cloud semantic segmentation aims to accurately segment "unseen" new categories in point cloud scenes using limited labeled data. However, pretraining-based methods not only introduce excessive time overhead but also overlook the local structure representation among irregular point clouds. To address these issues, we propose a pretraining-free local structure fitting network for few-shot point cloud semantic segmentation, named \textbf{TaylorSeg}. Specifically, inspired by Taylor series, we treat the local structure representation of irregular point clouds as a polynomial fitting problem and propose a novel local structure fitting convolution, called \textbf{TaylorConv}. This convolution learns the low-order basic information and high-order refined information of point clouds from explicit encoding of local geometric structures. Then, using TaylorConv as the basic component, we construct two variant of TaylorSeg: a non-parametric \textbf{TaylorSeg-NN} and a parametric \textbf{TaylorSeg-PN}. The former can achieve performance comparable to existing parametric models without pretraining. For the latter, we equip it with an \textbf{A}daptive \textbf{P}ush-\textbf{P}ull (\textbf{APP}) module to mitigate the feature distribution differences between the query set and the support set. Extensive experiments validate the effectiveness of the proposed method. Notably, under the 2-way 1-shot setting, TaylorSeg-PN achieves improvements of +2.28\% and +4.37\% mIoU on the S3DIS and ScanNet datasets respectively, compared to the previous state-of-the-art methods.
\end{abstract}

 \begin{links}
     \link{Code}{https://github.com/changshuowang/TaylorSeg}
 \end{links}

\begin{figure}[t]
	\centering
	\includegraphics[width=0.95\columnwidth]{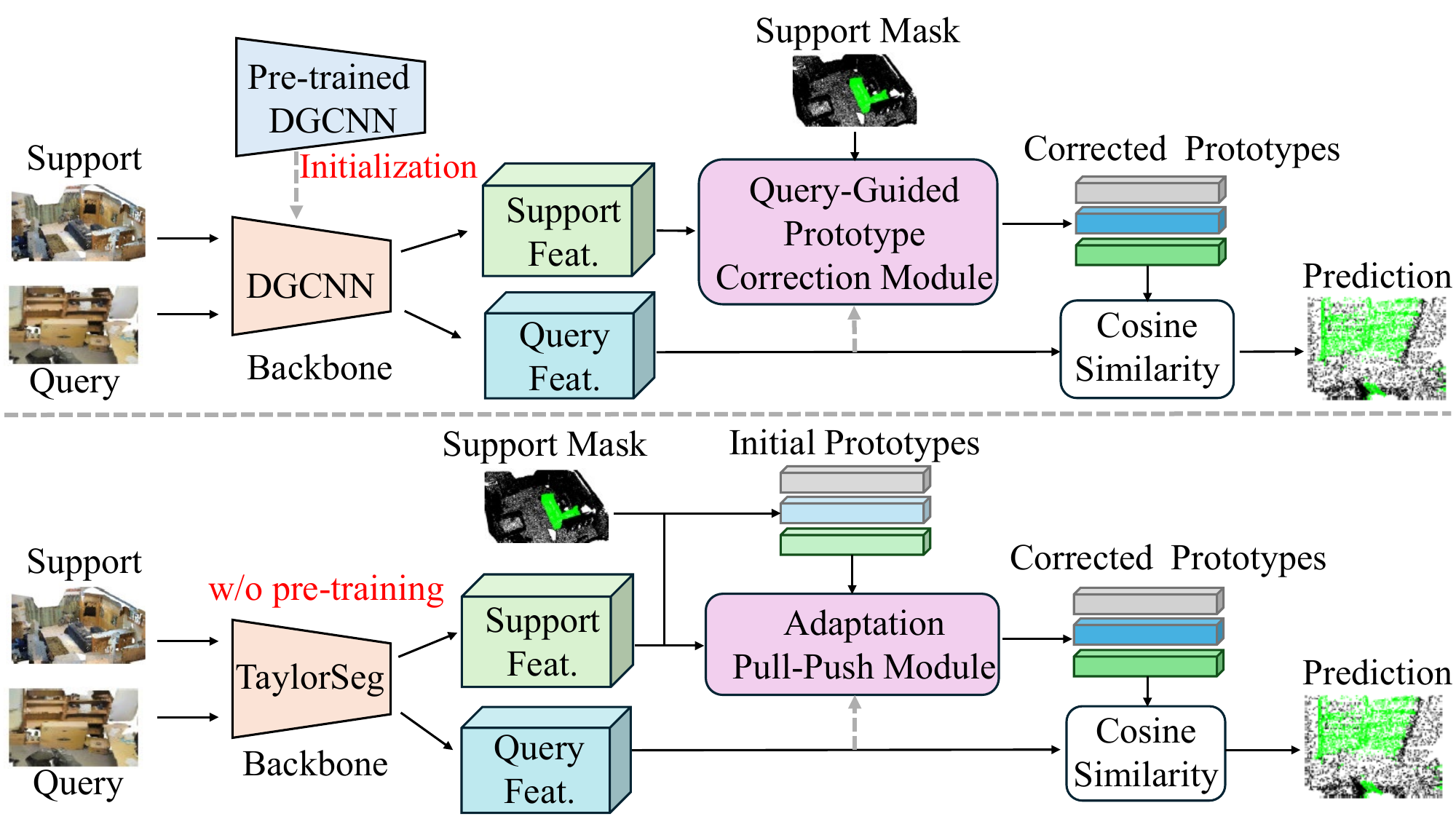}
	\caption{\textbf{Top}: Most existing methods are based on fine-tuning a pre-trained DGCNN, followed by using query features to guide and align the prototype features. This strategy is not only time-consuming but also overlooks the importance of local structure representation. \textbf{Bottom}: We propose a new backbone for point cloud tasks that requires no pre-training and possesses strong local structure representation capabilities. Additionally, we design an APP module that effectively aligns query features with prototype features.}
	\label{f1}
\end{figure}

\section{Introduction}  

Point cloud semantic segmentation \cite{zhang2023deep, zhang2024pointgt, zhang2024deformation} is an essential task in numerous computer vision applications \cite{wang2021brief, wang2025learning, wang2025point, fang2022multi,fang2024multi,fang2025adaptive,liu2024unsupervised,liu2023exploring,liu2023hypotheses,liu2023conditional}, such as autonomous driving \cite{wang2025destination,fei2024enhancing,fei2024vitron, yu2025eds}, robotics \cite{soori2023artificial,hu2024towards,hu2023transformer}, and augmented reality \cite{sereno2020collaborative, zhao2024harmonizing}. This task involves accurately assigning each point in the three-dimensional space to a specific semantic category. Recent advances in deep learning have led to significant improvements in many tasks \cite{fei2024video,fei2024dysen, fang2024uncertainty, fang2025rethinking,yu2024pedestrian}, particularly in tasks involving complex scene understanding \cite{tang2022optimal, tang2024textsquare, tang2022few, feng2023docpedia}. However, constructing large-scale, accurately annotated datasets requires enormous time and human resources, limiting the widespread applicability of these methods in real-world scenarios \cite{fang2023you,fang2024not,fang2024fewer,ning2023occluded,ning2023pedestrian}. Moreover, existing methods often struggle to maintain good generalization when faced with unseen new categories. Therefore, achieving efficient and accurate point cloud semantic segmentation with limited annotated data while endowing models with the ability to recognize unseen categories has become a critical research challenge.

To address these challenges, few-shot learning strategies \cite{snell2017prototypical} have gained widespread attention as an effective method to alleviate the issue of data scarcity. The core idea of few-shot learning is to achieve effective segmentation of new categories with a small number of labeled samples. Zhao et al. \cite{zhao2021few} introduced attMPTI, applying this concept to the few-shot point cloud semantic segmentation. This method first pretrains DGCNN \cite{wang2019dynamic} and then uses it as the encoder to extract feature embeddings from the support and query sets. Finally, each category in the support set is represented by multiple prototypes, utilizing a transductive label propagation method to associate the labeled prototypes with the unlabeled query points to recognize unseen categories. Following this approach, as shown in Fig.\ref{f1}(a), subsequent approaches \cite{mao2022bidirectional, zhu2023cross, zhang2023few} have focused on prototype construction and reducing the feature distribution gap between the support set and the query set. However, these methods suffer from three significant issues: (1) Pretraining DGCNN on seen categories introduces severe class bias when evaluating on unseen categories, reducing the model's generalization; (2) The pretraining phase increases time and resource costs; (3) Using DGCNN as the backbone limits the effective extraction of local structural information, affecting downstream prototype construction. Although the recently proposed Seg-NN \cite{zhu2024no} avoids the burden of pretraining, its local structural representation module lacks learnability.

To address the aforementioned issues, we propose a pretraining-free local structure fitting network, TaylorSeg, for few-shot point clouds semantic segmentation. First, the local structure feature extractor is the core component of the point cloud encoder, significantly impacting the prototype representation of the support set under scarce sample conditions. Inspired by Taylor series \cite{rudin1964principles}, we design a novel local structure fitting convolution, called TaylorConv. In this convolution, we view the local structure representation as a polynomial fitting problem to capture subtle changes in local geometric information more accurately. Specifically, the lower-order terms of this convolution are used to fit the flat regions of the local structure, which typically contain the basic shapes and overall trends of the point cloud. Higher-order terms are used to fit the edges and detailed parts, capturing complex variations and fine features within the local structure. Based on this convolution, we construct two variants of TaylorSeg: the non-parametric TaylorSeg-NN and the parametric TaylorSeg-PN. TaylorSeg-NN achieves performance equal to or better than existing parametric models without pretraining and with zero parameters. As an upgrade of TaylorSeg-NN, we parameterize TaylorConv to enhance the ability to represent local structures, thereby constructing TaylorSeg-PN. Additionally, to further enhance the capability of TaylorSeg-PN, we equip it with a learnable Adaptive Push-Pull (APP) module. This module learns common features between the support set and query set while distancing irrelevant features to bring their feature distributions closer together, thereby increasing the model's generalization ability.

In summary, our contributions are as follows:

\begin{itemize}
	\item We propose a pretraining-free local structure fitting network, TaylorSeg, for few-shot point cloud semantic segmentation. It exhibits strong generalization ability on unseen categories.
	
	\item Inspired by the Taylor series, we introduce TaylorConv, a novel operator that models local representation as a polynomial fitting problem, accurately capturing implicit shape information in local geometry.
	
	\item We design a new adaptive push-pull module, which can learn specific and common features between the support set and query set to bring their feature distributions closer.
	
	\item We conducted extensive experiments on S3DIS and ScanNet datasets, achieving state-of-the-art results with fewer parameters and faster efficiency.
\end{itemize}

\section{Relatation Works}

\subsection{3D Point Cloud Semantic Segmentation}
3D point cloud semantic segmentation \cite{zhou2024dynamic,liang2024pointmamba,liang2024pointgst, he2024refmask3d, he2025segpoint} aims to assign a specific label to each individual point in a given point cloud, and it has been extensively studied in the field of computer vision \cite{fang2024your, fang2023annotations, fang2023hierarchical, fang2021unbalanced, fang2021animc, zhao2024multi,xiong2024rethinking,cai2025imperceptible,liu2024towards,tang2025simplification,tang2024reparameterization,lei2025exploring,yang2025eood,liu2024pandora,zhang2025manipulating,zhang2025monoAttack,wang2025looking}. The pioneering work, PointNet \cite{qi2017pointnet}, introduced an end-to-end symmetric MLP network for segmenting raw point clouds. However, due to PointNet's lack of local structure information, PointNet++ \cite{qi2017pointnet} was proposed, which groups point clouds into different local neighborhoods using Farthest Point Sampling (FPS) and aggregates features within each local neighborhood. Nevertheless, this method still does not model the spatial relationships within local neighborhoods. Subsequent works \cite{wang2022learning, wang20233d,wangchangshuo20223d,jiang2023lttpoint} have primarily focused on methods based on point-wise Multi-Layer Perceptron (MLP), convolution operations, and attention mechanisms. Recent works \cite{wang2024gpsformer,park2023self} have made significant improvements by designing various Transformer structures to learn long-range dependencies in point clouds. Despite these advancements, these methods still require large amounts of labeled data, which is impractical in real-world applications. Furthermore, these methods struggle to recognize unseen categories. 
\begin{figure*}[t]
	\centering
	\includegraphics[width=0.85\textwidth]{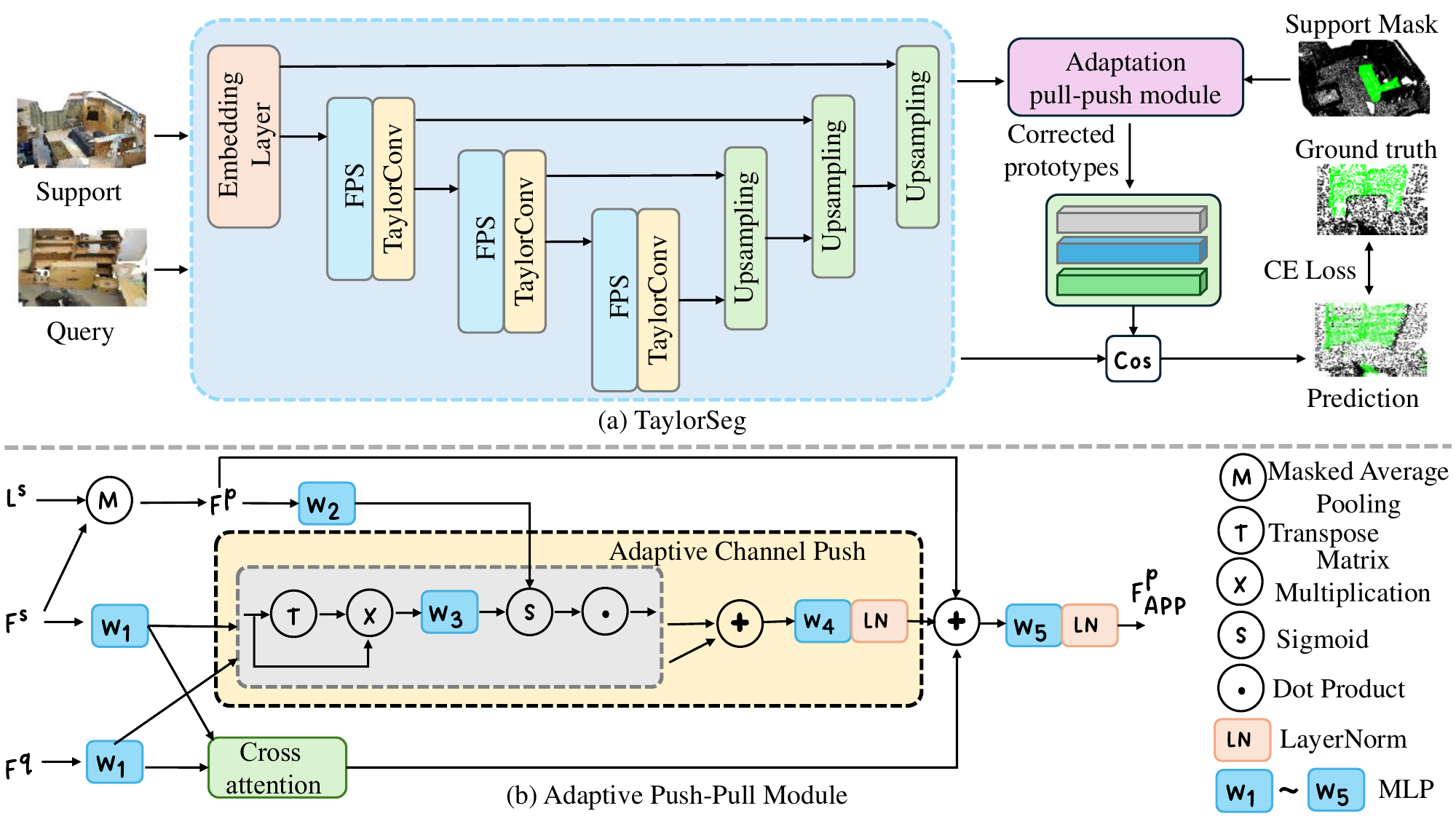}
	\caption{(a) The overall architecture of TaylorSeg. It centers around TayConv, a locally feature extraction module inspired by the Taylor series. TayConv forms the Taylor Block when combined with the FPS operation, and stacking these blocks along with upsampling operations constitutes our backbone network. TaylorSeg has two variants: for TaylorSeg-NN, the APP module is replaced with masked average pooling, allowing direct testing without training. For TaylorSeg-PN, both the TaylorConv and APP modules are optimized during training before testing. (b) The data flow diagram of the APP module. It is primarily designed for TaylorSeg-PN. It is not only parameter-efficient but also significantly reduces the feature distribution discrepancy between the query and support sets. Additionally, it is plug-and-play, making it compatible with many few-shot methods.}
	\label{f2}
\end{figure*}
\vspace{-10pt}

\subsection{Few-shot Point Cloud Semantic Segmentation}

Few-shot point cloud semantic segmentation \cite{an2024rethinking, liu2024dynamic, ning2023boosting} is the task of accurately predicting the categories of unknown point clouds using a limited number of training samples. As a pioneering work, AttMPTI \cite{zhao2021few} introduced few-shot learning of image processing tasks \cite{fang2020v,fang2020double,tang2022you, tang2023character,liu2023spts, ning2024enhancement} into point cloud semantic segmentation by first pretraining DGCNN \cite{wang2019dynamic} and then using it as the backbone for the encoder. It predicts the category of each instance in the query set by representing each class in the support set with multiple prototypes. BFG \cite{mao2022bidirectional} enhances the global perception ability of points by embedding global awareness bi-directionally into local points through the similarity measurement between the point features and the prototype features. 2CBR \cite{zhu2023cross} proposed cross-class rectification to alleviate the domain gap between the query set and the support set. PAP-FZ3D \cite{he2023prototype} designed the QGPA module to reduce the feature distribution differences between the query set and the support set while introducing word embeddings into zero/few-shot point cloud semantic segmentation tasks. However, these tasks increase time costs and resource consumption due to their reliance on pretraining strategies, and the DGCNN they depend on cannot sufficiently learn shape information from the local structures of point clouds. Although the recent Seg-NN \cite{zhu2024no} eliminates the need for pretraining, its local structure learner lacks the ability to learn and encode local structure information. To solve these issues, we propose the novel TaylorSeg, inspired by the Taylor series \cite{rudin1964principles}, which not only eliminates the need for pretraining but also fully learns local structural information.

\section{Methods}

In this section, we first review the problem definition of few-shot tasks; then, we introduce the concept of the Taylor series; next, we present TaylorConv. Finally, we explain how to construct TaylorSeg which is shown in Fig.\ref{f2}.

\subsection{Problem Definition}  
We divide all categories of the dataset into two classes: visible classes $C_{seen}$ and invisible classes $C_{unseen}$, satisfying $C_{seen} \cap C_{unseen} = \emptyset$. The sample set constructed from visible classes $C_{seen}$ serves as the training set, while the sample set from invisible classes $C_{unseen}$ serves as the test set. Following \cite{he2023prototype}, we also adopt the episodic paradigm. Specifically, each few-shot task (a.k.a. an episode) instantiates each training/testing set as an N-way K-shot segmentation learning task.

The support set can be represented as $S = \{(P_S^{(1,k)}, M_S^{(1,k)}), \ldots, (P_S^{(N,k)}, M_S^{(N,k)})\}_{k=1}^K$, which consists of $K$ annotated $P_S^{(n,k)}$ and their corresponding binary masks $M_S^{(n,k)} \in \mathbb{R}^{N_s \times 1}$, where $N_s$ represents the number of points in support sample. The query set $Q = \{(P_q^i, M_q^i)\}_{i=1}^T$, where $T$ is the number of query samples, and $M_q^i \in \mathbb{R}^{N_q \times 1}$ is the ground truth of the query point cloud, which is only available during training. $N_q$ represents the number of points in the query sample.

Typically, each point cloud $P \in \mathbb{R}^{N_p \times (3+C)}$ consists of $N_p$ points, where each point is composed of 3-dimensional coordinates and $C$-dimensional additional information (color, normal vector, etc.). In each episode, by inputting $(P_S^{(1,k)}, M_S^{(1,k)}, P_q^i)$, we predict the category of the query point cloud $P_q^i$ and calculate the cross-entropy loss with the ground truth $M_q^i$ to obtain an optimized model.

\subsection{Taylor Series} 
Taylor series \cite{rudin1964principles} has many applications in various fields. It can be used to approximate the value of complex functions and can be expressed as:
\begin{equation}
	f(x) = f(a) + \sum_{n=1}^{\infty} \frac{f^{(n)}(a)}{n!}(x - a)^n, \quad |x - a| < \epsilon ,
\end{equation}
where $a$ is a constant, $\epsilon$ is infinitesimal, and $f^{(n)}(a)$ represents the $n$-th derivative of $a$. To simplify the calculation, we will approximate the Taylor series with a low-frequency term and a high-frequency term. The specific formula is shown below:
\begin{equation}
	\label{e111}
	f(x) = f(a) + \sum_{n=1}^{\infty} a_n (x - a)^n, \quad |x - a| < \epsilon,
\end{equation}
where $a_n = \frac{f^{(n)}(a)}{n!}$. From the above, we can see that the Taylor series only needs to learn the local information near the constant $a$, and also needs to learn the relative information of $x$ and $a$. Therefore, we can easily introduce this property into the local structure representation of point cloud.

\subsection{Local Structure Fitting Convolution}
Suppose the coordinates and features of the $i$-th point of a point cloud can be represented as $p_i \in \mathbb{R}^{1\times3}$ and $f_i \in \mathbb{R}^{1\times C}$, where $C$ denotes the number of channels. Then, with $p_i$ as the center point, the local feature aggregation can be formalized as:
\begin{equation}
	f_i' = \mathcal{A}(\{\mathcal{M}(p_i, p_j) \cdot \mathcal{T}(f_i, f_j) | p_j \in \mathcal{N}(p_i)\}).
\end{equation}
Here, $\mathcal{A}$ is the aggregation function, typically max pooling. $\mathcal{M}$ and $\mathcal{T}$ are mapping functions, usually MLPs. $\mathcal{N}(p_i)$ is the local neighborhood of $p_i$, and $p_j$ is $p_i$'s neighboring point. This formula simply maps the coordinates and features to a high-dimensional space, which cannot accurately model the local structure of the point cloud. This coarse-grained local structure learning is far from sufficient for tasks under sample-scarce conditions.

Due to the complexity of the point cloud scene, accurately representing local structural information plays a crucial role in extracting discriminative features and effective representation of the support set. Taylor series \cite{rudin1964principles} can be used to approximate the value of complex functions. It not only contains overall information but also includes high-frequency information. Inspired by this, we propose a local structure fitting convolution, called TaylorConv, to refine the analysis of the local structure and detail information of the point cloud. Specifically, TaylorConv consists of two parts: Low-order Convolution (LoConv) and High-order Convolution (HiConv). The former learns the overall structural information (low-frequency information) $f_i^L$ of the local structure through max pooling of all neighboring points. The latter mainly learns the refined detail information (high-frequency) $f_i^H$ from the relative features of neighboring points. Therefore, TaylorConv can be formalized as:
\begin{equation}
	%\footnotesize
	g \approx f_i^L + f_i^H = \mathcal{A}(\{\phi(f_j)\}_{j=1}^L) + \mathcal{A}(\{\mathcal{T}(f_i, f_j)\}_{j=1}^L),
\end{equation}
where $g$ is the output feature, $\phi$ represents the feature mapping function, $L$ represents the number of neighboring points. $\mathcal{T}(f_i, f_j) = \left(\frac{w_j\cdot(f_j-f_i)}{|w_j\cdot(f_j-f_i)|}\right)^s \cdot |w_j \cdot (f_j - f_i)|^p$ is a new type of isometry function we designed, called high-order kernel function. Here, $|\cdot|$ represents the absolute value of vector elements, $s \in \{0,1\}$, $p$ is an integer parameter. When $s=1$, $p=1$, $\mathcal{T}(f_i, f_j)$ degenerates to the traditional Affine Basis Function (ABF \cite{rosenblatt1958perceptron}) (see Eq. \ref{e1}); when $s=0$, $p=2$, $\mathcal{T}(f_i, f_j)$ degenerates to the Radial Basis Function (RBF \cite{moody1989fast}) (see \ref{e2}). Therefore, the proposed high-order kernel function has strong expressiveness.
\begin{equation}
	\label{e1}
	\mathcal{T}=\left(\frac{w_j \cdot\left(f_j-0\right)}{\left|w_j \cdot\left(f_j-0\right)\right|}\right)^1 \cdot\left|w_j \cdot\left(f_j-0\right)\right|^1=w_j \cdot f_j,
\end{equation}
\begin{equation}
	\label{e2}
	\mathcal{T}=\left(\frac{w_j \cdot\left(f_j-f_i\right)}{\left|w_j \cdot\left(f_j-f_i\right)\right|}\right)^0 \cdot\left|w_j \cdot\left(f_j-f_i\right)\right|^2.
\end{equation}

\subsection{Local Structure Fitting Network}
As shown in Fig.\ref{f2}(a), we have developed two variants of TaylorSeg based on TaylorConv: the parameter-free TaylorSeg-NN and the parameterized TaylorSeg-PN. Both variants do not require pre-training, with TaylorSeg-NN in particular achieving desirable results without any training. The encoder of TaylorSeg is composed of stacked Taylor Blocks, which are constructed using FPS and TaylorConv. The decoder restores the resolution of the point cloud through upsampling. Between the encoder and decoder, we have implemented skip connections similar to the U-Net.

\subsubsection{TaylorSeg-NN.} To realize the parameter-free TaylorSeg-NN, we borrowed from Point-NN \cite{zhang2023starting} and used trigonometric PEs for feature mapping. Suppose there is a point $x \in \mathbb{R}^{1\times d}$, where $d$ represents the dimension, then Trigonometric PEs can be expressed as $E(\cdot)$:
\begin{equation}
	E(x; u) = [\sin(2\pi ux), \cos(2\pi ux)] \in \mathbb{R}^{1\times 6d},
\end{equation}
where $u = [u_1, ..., u_d], u_d = 6^i, i = 1, ..., d$, $\theta$ is a super parameter. Since TaylorSeg-NN has no learnable parameters and only simple numerical transformations, in order to define the local structure information, we use 0-order and 1-order features. Thus, TaylorConv can be expressed as:
\begin{equation}
	g \approx \mathcal{A}\left(\left\{f_j\right\}_{j=1}^K\right) + \mathcal{A}\left(\left\{w_j \cdot f_j\right\}_{j=1}^K\right),
\end{equation}
where $f_j = (E(p_j; u) + E(c_j; u) + E(f_j^g; u))/3$, $w_j = \cos(2\pi \cdot E([p_i, p_j, p_j - p_i]; u))$. $c_j$ is the color information of the $j$-th point, $f_j^g$ is the first layer of the point's feature. Here we have introduced the geometric coordinate information into the representation of local structural information, so TaylorSeg-NN can effectively model the spatial distribution relationship of objects.

\subsubsection{TaylorSeg-PN} 
To further enhance the expressive ability of TaylorSeg-NN, we obtain $w_j = MLP(p_i, p_j, p_j - p_i)$ of HiConv through a learnable MLP. To further improve TaylorSeg-PN's ability to generalize to unseen categories, we propose a Adaptive Push-Pull (APP) module. It consists of two parts: Adaptive Channel Push (ACP) and Cross Attention Pull (CAP). First, we perform local max pooling and mapping along the point dimension of the support feature $F^s \in \mathbb{R}^{M \times C}$ and query feature $F^q \in \mathbb{R}^{M \times C}$ separately to learn the statistical characteristics of each channel. We also further map the prototype feature $F^p$ to increase its flexibility. The specific formulas are as follows:
\begin{equation}
	{F}^s = W_1 \cdot MaxPooling(F^s) \in \mathbb{R}^{M' \times C},
\end{equation}
\begin{equation}
	{F}^q = W_1 \cdot MaxPooling(F^q) \in \mathbb{R}^{M' \times C},
\end{equation}
\begin{equation}
	\bar{F}^p = W_2 \cdot F^p \in \mathbb{R}^{K \times C},
\end{equation}
where $W_1$ and $W_2$ are learnable fully connected layers.

\textbf{Adaptive Channel Push.} In this stage, we focus on learning the correlation between the support feature and query feature for each channel. Then we push the original features to their respective feature distribution spaces. The specific formula is as follows:
\begin{equation}
	G^s = (F^s)^T F^s \in \mathbb{R}^{C \times C},
\end{equation}
\begin{equation}
	A^s = Sigmoid(W_3 \cdot G^s) \in \mathbb{R}^{1 \times C}.
\end{equation}

Similarly, we can obtain the channel attention weight $A^q$ for the query set. Finally, we push the prototype to different distribution spaces according to the channel attention weights of the support set and query set respectively, and fuse these two prototype features. The formula is as follows:
\begin{equation}
	F_{ACP}^p = LN(W_4(A^s \cdot \bar{F}^p + A^q \cdot \bar{F}^p)),
\end{equation}
where $W_4$ is a learnable fully connected layer, $LN$ represents the layer normalization layer.

\textbf{Cross Attention Pull.} Due to the complexity of point clouds and the scarcity of samples, simply fusing prototype features in the spatial distribution of support features and query features cannot fully narrow the gap between query features and prototype features. Therefore, we learn the common features of the two domains through cross-attention. The specific formula is as follows:
\begin{equation}
	A_{cross} = (F^q)^T F^s \in \mathbb{R}^{C \times C},
\end{equation}
\begin{equation}
	F_{CAP}^p = Softmax(A_{cross}) \cdot (F^p)^T.
\end{equation}

Finally, the updated prototype feature output by the APP module is: $F_{APP}^p = LN(W_5(F_{ACP}^p + F_{CAP}^p) + F^p)$. Here, $W_5$ is a learnable fully connected layer. For N-way-K-shot tasks, we average all K-shot prototype features as the final prototype feature.

During the training process, we alternately optimize TaylorConv and the APP module. In the testing phase, we pair the prototype features output by the APP module with the query set features to perform similarity matching, thereby achieving segmentation of unseen points.

\begin{table*}[t]
	\centering

	\renewcommand\arraystretch{1.2}
	\resizebox{1.95\columnwidth}{!}{
		\begin{tabular}{l|c||ccc|ccc||ccc|ccc}
			\hline 
			\multirow{3}{*}{\textbf{Method}} & \multirow{3}{*}{\textbf{Param.}} & \multicolumn{6}{c||}{\textbf{Two-way}} & \multicolumn{6}{c}{\textbf{Three-way}} \\
			\cline{3-14}
			& & \multicolumn{3}{c|}{\textbf{One-shot}} & \multicolumn{3}{c||}{\textbf{Five-shot}} & \multicolumn{3}{c|}{\textbf{One-shot}} & \multicolumn{3}{c}{\textbf{Five-shot}} \\
			\cline{3-14}
			& & $S_0$ & $S_1$ & $Avg$ & $S_0$ & $S_1$ & $Avg$ & $S_0$ & $S_1$ & $Avg$ & $S_0$ & $S_1$ & $Avg$ \\
			\hline
			\rowcolor{green!5} Point-NN & 0.00 M & 42.12 & 42.62 & 42.37 & 51.91 & 49.35 & 50.63 & 38.00 & 36.21 & 37.10 & 45.91 & 43.44 & 44.67 \\
			\rowcolor{green!5} Seg-NN & 0.00 M & 49.45 & 49.60 & 49.53 & 59.40 & 61.48 & 60.44 & 39.06 & 40.10 & 39.58 & 50.14 & 51.33 & 50.74 \\
			\rowcolor{green!5} \textbf{TaylorSeg-NN} & 0.00 M & 52.30 & 50.51 & 51.41 & 62.29 & 62.30 & 62.30 & 40.76 & 40.24 & 40.50 & 51.95 & 51.52 & 51.44 \\
			\rowcolor{green!5} \emph{Improvement} & - & \textcolor{blue}{+2.85} & \textcolor{blue}{+0.91} & \textcolor{blue}{+1.88} & \textcolor{blue}{+2.89} & \textcolor{blue}{+0.82} & \textcolor{blue}{+1.86} & \textcolor{blue}{+1.70} & \textcolor{blue}{+0.14} & \textcolor{blue}{+0.92} & \textcolor{blue}{+1.81} & \textcolor{blue}{+0.19} & \textcolor{blue}{+1.0} \\
			\hline
			DGCNN & 0.62 M & 36.34 & 38.79 & 37.57 & 56.49 & 56.99 & 56.74 & 30.05 & 32.19 & 31.12 & 46.88 & 47.57 & 47.23 \\
			ProtoNet & 0.27 M & 48.39 & 49.98 & 49.19 & 57.34 & 63.22 & 60.28 & 40.81 & 45.07 & 42.94 & 49.05 & 53.42 & 51.24 \\
			MPTI & 0.29 M & 52.27 & 51.48 & 51.88 & 58.93 & 60.56 & 59.75 & 44.27 & 46.92 & 45.60 & 51.74 & 48.57 & 50.16 \\
			AttMPTI & 0.37 M & 53.77 & 55.94 & 54.86 & 61.67 & 67.02 & 64.35 & 45.18 & 49.27 & 47.23 & 54.92 & 56.79 & 55.86 \\
			BFG & - & 55.60 & 55.98 & 55.79 & 63.71 & 66.62 & 65.17 & 46.18 & 48.36 & 47.27 & 55.05 & 57.80 & 56.43 \\
			2CBR & 0.35 M & 55.89 & 61.99 & 58.94 & 63.55 & 67.51 & 65.53 & 46.51 & 53.91 & 50.21 & 55.51 & 58.07 & 56.79 \\
			PAP3D & 2.45 M & 59.45 & 66.08 & 62.76 & 65.40 & 70.30 & 67.85 & 48.99 & 56.57 & 52.78 & 61.27 & 60.81 & 61.04 \\
			Seg-PN & 0.24 M & 64.84 & 67.98 & 66.41 & 67.63 & 71.48 & 69.36 & 60.12 & 63.22 & 61.67 & 62.58 & 64.53 & 63.56 \\
			\hline
			\textbf{TaylorSeg-PN} & 0.27 M & \textbf{67.12} & \textbf{71.11} & \textbf{69.12} & \textbf{70.44} & \textbf{72.23} & \textbf{71.34} & \textbf{60.28} & \textbf{65.70} & \textbf{63.00} & \textbf{62.78} & \textbf{67.06} & \textbf{64.33} \\
			\emph{Improvement} & - & \textcolor{blue}{+2.28} & \textcolor{blue}{+3.13} & \textcolor{blue}{+2.71} & \textcolor{blue}{+2.81} & \textcolor{blue}{+0.75} & \textcolor{blue}{+1.98} & \textcolor{blue}{+0.16} & \textcolor{blue}{+2.48} & \textcolor{blue}{+1.32} & \textcolor{blue}{+0.20} & \textcolor{blue}{+2.53} & \textcolor{blue}{+1.37} \\
			\hline
		\end{tabular}
	}
	
		\caption{\textbf{Few-shot Results (\%) on S3DIS.} $S_i$ denotes the split $i$ is used for testing, and \emph{Avg} is their average mIoU. The shaded rows represent non-parametric methods. 'Param.' represents the total number of learnable parameters of each method.}
		
		\label{t1}
\end{table*}

\section{Experiments} 
In this section, we first introduce the datasets used and the experimental details. Then, we report the experimental results of TaylorSeg on the S3DIS and ScanNet datasets. Finally, we validate the effectiveness of our method through ablation experiments.

\subsection{Datasets and Evaluation Metrics}
\textbf{Datasets:} S3DIS \cite{armeni20163d} is a dataset of 3D RGB point clouds collected from 272 rooms across 6 indoor environments. Each point is annotated with one of 13 semantic labels (12 semantic classes plus clutter). The ScanNet \cite{dai2017scannet} dataset contains a total of 1513 scanned scenes. All points, except for unannotated spaces, are annotated with 20 semantic classes.

Following \cite{zhao2021few}, we divide each point cloud scene into $1 \mathrm{~m} \times 1 \mathrm{~m}$ blocks, randomly sampling 2048 points from each block. The S3DIS and ScanNet datasets are divided into 7547 and 36350 blocks respectively. Simultaneously, each dataset is divided into two non-overlapping subsets of categories, denoted as $S_0$ and $S_1$. When one subset is designated as the test set, the other subset is designated as the training set.

\textbf{Evaluation Metric:} We choose mIoU (Mean Intersection over Union), which is widely used in point cloud segmentation, as the performance evaluation metric.

\subsection{Implementation Details}
We conducted all experiments by pytorch framework and using one GeForce RTX 4090 GPU. For TaylorSeg-NN, its encoder was frozen in all experiments, and the local neighborhood in TaylorConv was determined by FPS, with each local neighborhood using 16 neighboring points from K-NN. In the initial layer of trigonometric PEs, we set the frequency parameter to 20, and $\theta$ to 30. Since TaylorSeg-NN doesn't require pre-training, its output doesn't need to go through APP, only requiring the decoder's features.

For TaylorSeg-PN, we set the HiConv to be learnable, while passing TaylorSeg-PN's output through the APP module to obtain the final prototype features. In APP, we set the max pooling stride to 32. During training, we use the AdamW optimizer ($\beta_1 = 0.9$, $\beta_2 = 0.999$) to update TaylorSeg-PN's TaylorConv and APP module. The initial learning rate is set to 0.001, halving every 7,000 iterations. In episodic training, each batch contains 1 episode, which includes one support set and one query set. The support set randomly selects N-way-K-shot, and the query set randomly selects N unseen samples.

\subsection{Comparison with State-of-the-Art Methods}
To evaluate our method, we compared it with both parameter-free methods (Point-NN \cite{zhang2023starting}, Seg-NN \cite{zhu2024no}) and parameterized methods (DGCNN \cite{wang2019dynamic}, ProtoNet \cite{garcia2017few}, MPTI \cite{zhao2021few}, AttMPTI \cite{zhao2021few}, BFG \cite{mao2022bidirectional}, 2CBR \cite{zhu2023cross}, PAP3D \cite{he2023prototype}, Seg-PN \cite{zhu2024no}).

\textbf{Results analysis on the S3DIS dataset.} As shown in Table \ref{t1}, for non-parametric methods, we compared TaylorSeg-NN against Point-NN \cite{zhang2023starting} and Seg-NN \cite{zhu2024no}. TaylorSeg-NN exhibited noticeable improvements across all settings, achieving +1.88\% average mIoU increase in the 2-way 1-shot setting. This results suggest that TaylorSeg's local structure fitting strategy is more effective in capturing subtle geometric features, even in scenarios with minimal training data. When comparing TaylorSeg-PN with its Seg-PN \cite{zhang2023starting}, we observed a remarkable increase of +2.71\% in average mIoU across the 2-way 1-shot settings. TaylorSeg-PN consistently outperformed other methods like PAP3D \cite{he2023prototype}, demonstrating its superiority in few-shot learning tasks by better mitigating the domain gap between seen and unseen classes.

\textbf{Results analysis on the ScanNet dataset.} As shown in Table \ref{t2}, similar trends were observed on the ScanNet dataset. TaylorSeg-NN outperformed Point-NN and Seg-NN, with an average mIoU improvement of +1.31\% in the 2-way 1-shot setting. This result indicates TaylorSeg's robust generalization capability across different datasets. For parametric models, TaylorSeg-PN delivered a significant +5.40\% improvement in average mIoU in the 2-way 1-shot scenario. Compared to the previous SOTA method, TaylorSeg-PN achieved better performance while utilizing similar parameters, highlighting its efficiency and effectiveness in leveraging limited labeled data.

\begin{table}[t]
	\centering

	\renewcommand\arraystretch{1.2}
	\resizebox{0.95\columnwidth}{!}{ 
		
		\begin{tabular}{l|ccc|ccc}
			\hline \multirow{2}{*}{ Setting } & \multicolumn{3}{c|}{ Two Way } & \multicolumn{3}{c}{ Three Way } \\
			\cline { 2 - 7 } & $S_0$ & $S_1$ & $A v g$ & $S_0$ & $S_1$ & $A v g$ \\
			\hline 
			ABF & 66.30 & 68.69 & 67.50 & 60.11 & 64.85 & 62.48 \\
			RBF & 56.46 & 62.10 & 59.28 & 59.89 & 57.41  & 58.65 \\
			s=0, $\mathrm{p}$ learnable & 65.97 & 67.96 & 66.97 & 59.59 & 64.85 & 62.22 \\
			s=1, $\mathrm{p}$ learnable & \textbf{67.12} & \textbf{71.11} & \textbf{69.12}  & \textbf{70.44} & \textbf{72.23} &  \textbf{71.34} \\
			\hline
		\end{tabular}
	}
	
	\caption{The Influence of HiConv’s Parameters on TaylorSeg-PN. We report
	the results (\%) under 2/3-way-1-shot settings on S3DIS datasets.}
	
	\label{t3}
	
\end{table}

\begin{table*}[t]
	\centering

	\renewcommand\arraystretch{1.2}
	\resizebox{1.95\columnwidth}{!}{
		\begin{tabular}{l|c||ccc|ccc||ccc|ccc}
			\hline 
			\multirow{3}{*}{\textbf{Method}} & \multirow{3}{*}{\textbf{Param.}} & \multicolumn{6}{c||}{\textbf{Two-way}} & \multicolumn{6}{c}{\textbf{Three-way}} \\
			\cline{3-14}
			& & \multicolumn{3}{c|}{\textbf{One-shot}} & \multicolumn{3}{c||}{\textbf{Five-shot}} & \multicolumn{3}{c|}{\textbf{One-shot}} & \multicolumn{3}{c}{\textbf{Five-shot}} \\
			\cline{3-14}
			& & $S_0$ & $S_1$ & Avg & $S_0$ & $S_1$ & Avg & $S_0$ & $S_1$ & Avg & $S_0$ & $S_1$ & Avg \\
			\hline
			\rowcolor{green!5} Point-NN & 0.00 M & 28.85 & 31.56 & 30.21 & 34.82 & 32.87 & 33.85 & 21.24 & 17.91 & 19.58 & 26.42 & 23.98 & 25.20 \\
			\rowcolor{green!5} Seg-NN & 0.00 M & 36.80 & 35.86 & 36.38 & 43.97 & 41.50 & 42.74 & 27.41 & 23.36 & 25.39 & 34.27 & 30.75 & 32.51 \\
			\rowcolor{green!5} \textbf{TaylorSeg-NN} & 0.00 M & 38.55 & 36.83 & 37.69 & 46.41 & 44.58 & 45.50 & 27.62 & 24.63 & 26.03 & 34.98 & 31.82 & 33.40 \\
			\rowcolor{green!5} \emph{Improvement} & - & \textcolor{blue}{+1.75} & \textcolor{blue}{+0.97} & \textcolor{blue}{+1.31} & \textcolor{blue}{+2.44} & \textcolor{blue}{+3.08} & \textcolor{blue}{+2.76} & \textcolor{blue}{+0.21} & \textcolor{blue}{+1.27} & \textcolor{blue}{+0.74} & \textcolor{blue}{+0.71} & \textcolor{blue}{+1.07} & \textcolor{blue}{+0.89} \\
			\hline
			DGCNN & 1.43 M & 31.55 & 28.94 & 30.25 & 42.71 & 37.24 & 39.98 & 23.99 & 19.10 & 21.55 & 34.93 & 28.10 & 31.52 \\
			ProtoNet & 0.27 M & 33.92 & 30.95 & 32.44 & 45.34 & 42.01 & 43.68 & 28.47 & 26.13 & 27.30 & 37.36 & 34.98 & 36.17 \\
			MPTI & 0.29 M & 39.27 & 36.14 & 37.71 & 46.90 & 43.59 & 45.25 & 29.96 & 27.26 & 28.61 & 38.14 & 34.36 & 36.25 \\
			AttMPTI & 0.37 M & 42.55 & 40.83 & 41.69 & 54.00 & 50.32 & 52.16 & 35.23 & 30.72 & 32.98 & 46.74 & 40.80 & 43.77 \\
			BFG & - & 42.15 & 40.52 & 41.34 & 51.23 & 49.39 & 50.31 & 34.12 & 31.98 & 33.05 & 46.25 & 41.38 & 43.82 \\
			2CBR & 0.35 M & 50.73 & 47.66 & 49.20 & 52.35 & 47.14 & 49.75 & 47.00 & 46.36 & 46.68 & 45.06 & 39.47 & 42.27 \\
			PAP3D & 2.45 M & 57.08 & 55.94 & 56.51 & 64.55 & 59.64 & 62.10 & 55.27 & 55.60 & 55.44 & 59.02 & 53.16 & 56.09 \\
			Seg-PN & 0.24 M & 63.15 & 64.32 & 63.74 & 67.08 & 69.05 & 68.07 & 61.80 & 65.34 & 63.57 & 62.94 & 68.26 & 65.60 \\
			\hline
			\textbf{TaylorSeg-PN} & 0.27 M & \textbf{67.52} & \textbf{70.75} & \textbf{69.14} & \textbf{68.39} & \textbf{71.55} & \textbf{69.97} & \textbf{63.60} & \textbf{67.55} & \textbf{65.58} & \textbf{66.98} & \textbf{69.78} & \textbf{68.38} \\
			\emph{Improvement} & - & \textcolor{blue}{+4.37} & \textcolor{blue}{+6.43} & \textcolor{blue}{+5.40} & \textcolor{blue}{+1.31} & \textcolor{blue}{+2.50} & \textcolor{blue}{+1.90} & \textcolor{blue}{+1.80} & \textcolor{blue}{+2.21} & \textcolor{blue}{+2.01} & \textcolor{blue}{+4.04} & \textcolor{blue}{+1.52} & \textcolor{blue}{+2.78} \\
			\hline
		\end{tabular}
	}
	\caption{\textbf{Few-shot Results (\%) on ScanNet.} $S_i$ denotes the split $i$ is used for testing, and \emph{Avg} is their average mIoU. The shaded rows represent non-parametric methods. 'Param.' represents the total number of learnable parameters of each method.}
	\label{t2}
\end{table*}

\subsection{Ablation Study}
\subsubsection{Ablation on HiConv.}

The results in Table \ref{t3} highlight the significant impact of the parameter \( s \) in the HiConv on TaylorSeg-PN's performance. Specifically, when \( s = 1 \), the model achieves the highest accuracy, with average scores of 69.12\% in 2-way and 71.34\% in 3-way settings. This demonstrates that HiConv, which effectively captures high-frequency details, is essential for improving the model's ability to learn complex local structures. Conversely, setting ABF or RBF results in reduced performance, indicating that a simpler form of the convolution is less effective. These findings confirm that the high-order kernel function plays a critical role in enhancing local feature representation.

\subsubsection{Ablation on APP Module.}
The results in Table \ref{t4} highlight the effectiveness of the APP module in enhancing TaylorSeg-PN's performance. The full APP module, which integrates both ACP and CAP, achieves the best results, with an average improvement of 18.07\% in 2-way and 29.57\% in 3-way settings compared to the baseline without APP. The CAP component alone also significantly boosts performance, but the combination of ACP and CAP maximizes the model's ability to generalize to unseen categories. This indicates that the APP module's approach of leveraging adaptive channel attention and cross-domain feature learning is highly effective in refining prototype features and improving segmentation accuracy under few-shot conditions.

\begin{table}[t]
	\centering
	
	\renewcommand\arraystretch{1.2}
	\resizebox{0.95\columnwidth}{!}{ 
		\begin{tabular}{l|ccc|ccc}
			\hline \multirow{2}{*}{ Setting } & \multicolumn{3}{c|}{ Two Way } & \multicolumn{3}{c}{ Three Way } \\
			\cline { 2 - 7 } & $S_0$ & $S_1$ & $A v g$ & $S_0$ & $S_1$ & $A v g$ \\
			\hline w/o & 49.42 & 52.67 & 51.05 & 40.66 & 42.87 & 41.77 \\
			ACP & 51.02 & 55.23 & 53.13 & 41.56 & 44.63 & 43.10 \\
			CAP & 64.64 & 68.54 & 66.59 & 61.45 & 64.12 & 62.79 \\
			APP & \textbf{67.12} & \textbf{71.11} & \textbf{69.12}  & \textbf{70.44} & \textbf{72.23} &  \textbf{71.34} \\
			\hline
		\end{tabular}
	}
	
	\caption{Ablation for the APP Module in TaylorSeg-PN. We report
		the results (\%) under 2/3-way-1-shot settings.}
		
	\label{t4}
\end{table}

\begin{table}[t]
	\centering

	\renewcommand\arraystretch{1.2}
	\resizebox{0.95\columnwidth}{!}{ 
		\begin{tabular}{cccccc}
			\hline \textbf{Method} & \textbf{mIoU} & \textbf{Param.} & \begin{tabular}{l} 
				\textbf{Pre-train} \\
				\textbf{Time}
			\end{tabular} & \begin{tabular}{l} 
				\textbf{Episodic} \\
				\textbf{Train}
			\end{tabular} & \begin{tabular}{l} 
				\textbf{Total} \\
				\textbf{Time}
			\end{tabular} \\ 
			\cmidrule(r){1-1} 
			\cmidrule(r){2-2} 
			\cmidrule(r){3-3} 
			\cmidrule(r){4-6} 
			DGCNN & 36.34 & 0.62 M & 4.0 h & 0.8 h & 4.8 h \\
			AttMPTI & 53.77 & 0.37 M & 4.0 h & 5.5 h & 9.5 h \\
			2CBR & 55.89 & 0.35 M & 6.0 h & 0.2 h & 6.2 h \\
			PAP3D & 59.45 & 2.45 M & 3.6 h & 1.1 h & 4.7 h \\
			Seg-NN & 49.45 & 0.00 M & 0.0 h & 0.0 h & 0.0 h \\
			Seg-PN & 64.84 & 0.24 M & 0.0 h & 0.5 h & 0.5 h \\
			\hline
			\textbf{TaylorSeg-NN} & 52.30 & 0.00 M & 0.0 h & 0.0 h & 0.0 h \\
			\textbf{TaylorSeg-PN} & 67.12 & 0.27 M & 0.0 h & 0.6 h & 0.6 h \\
			\hline
		\end{tabular}
	}
	
		\caption{Performance and efficiency comparison on S3DIS under 2-way-1-shot settings on the $S_0$ split. Training time and test speed (episodes/second) are measured on a single NVIDIA A6000 GPU.}
		
		\label{t7}
		
\end{table}

\subsubsection{Different Numbers of Encoder Layers}

\begin{figure}[t]
	\centering
	\includegraphics[width=0.95\columnwidth]{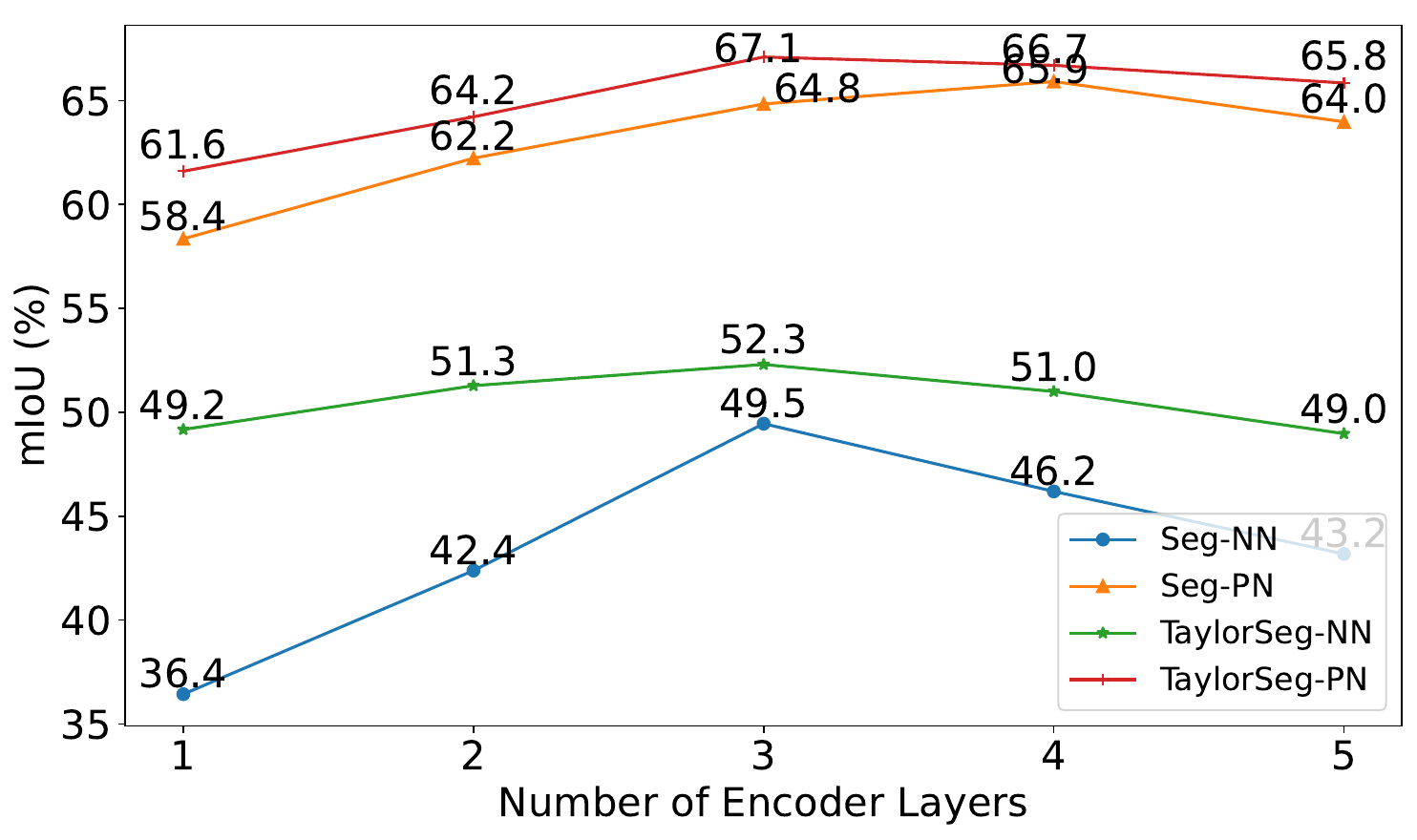}
	\caption{Ablation for Number of Encoder Layers in 2-way-1-shot setting on the S3DIS dataset.}
	\label{f3}
\end{figure}

Based on Fig. \ref{f3}, we observe that both TaylorSeg-NN and TaylorSeg-PN models show variation in performance with the number of encoder layers. TaylorSeg-NN peaks at three layers, balancing feature learning and model efficiency, while performance declines beyond this point due to potential overfitting. TaylorSeg-PN also achieves its best performance with three layers but declines more gradually, maintaining strong performance with four layers. Notably, TaylorSeg consistently outperforms Seg-NN across all configurations, highlighting its superior architecture for capturing local point cloud structures and improving segmentation accuracy.

\subsection{Computational Complexity}

In Table \ref{t7}, we compare the training time and efficiency of our method with other approaches. TaylorSeg-NN and Seg-NN achieve few-shot point cloud segmentation with minimal resources and time. Notably, TaylorSeg-NN surpasses Seg-NN in accuracy by 2.85\%. Additionally, when adjusted for equivalent computational cost, TaylorSeg-PN, with only a marginal increase of 0.03M parameters and 0.1h training time, outperforms Seg-PN by 2.28\% mIoU.

\section{Conclusion}

In this paper, we introduced TaylorSeg, a pretraining-free local structure fitting network for few-shot point cloud semantic segmentation. Our key contributions are TaylorConv and the Adaptive Push-Pull (APP) module. TaylorConv captures both low-order and high-order geometric details, while APP bridges the feature distribution gap between query sets and prototypes. TaylorSeg-NN achieves competitive performance without training, while TaylorSeg-PN, with learnable components, achieves state-of-the-art results on S3DIS and ScanNet datasets. Future work will focus on reducing TaylorConv's computational complexity, integrating textual information for few-shot tasks, and exploring TaylorSeg's potential in other 3D vision applications.

\section*{Acknowledgments}
This work was supported in part by the Ministry of Education, Singapore, under its Academic Research Fund Tier 1, under Grant RG78/21.

\bibliography{aaai25}

\begin{thebibliography}{89}
\providecommand{\natexlab}[1]{#1}

\bibitem[{An et~al.(2024)An, Sun, Liu, Liu, Wu, Wang, Van~Gool, and
  Belongie}]{an2024rethinking}
An, Z.; Sun, G.; Liu, Y.; Liu, F.; Wu, Z.; Wang, D.; Van~Gool, L.; and
  Belongie, S. 2024.
\newblock Rethinking Few-shot 3D Point Cloud Semantic Segmentation.
\newblock In \emph{CVPR}, 3996--4006.

\bibitem[{Armeni et~al.(2016)Armeni, Sener, Zamir, Jiang, Brilakis, Fischer,
  and Savarese}]{armeni20163d}
Armeni, I.; Sener, O.; Zamir, A.~R.; Jiang, H.; Brilakis, I.; Fischer, M.; and
  Savarese, S. 2016.
\newblock 3d semantic parsing of large-scale indoor spaces.
\newblock In \emph{CVPR}, 1534--1543.

\bibitem[{Cai et~al.(2025)Cai, Liu, Fang, Yu, Tang, and
  Zhou}]{cai2025imperceptible}
Cai, F.; Liu, D.; Fang, X.; Yu, J.; Tang, K.; and Zhou, P. 2025.
\newblock Imperceptible Beam-Sensitive Adversarial Attacks for LiDAR-based
  Object Detection in Autonomous Driving.
\newblock \emph{ICME}.

\bibitem[{Dai et~al.(2017)Dai, Chang, Savva, Halber, Funkhouser, and
  Nie{\ss}ner}]{dai2017scannet}
Dai, A.; Chang, A.~X.; Savva, M.; Halber, M.; Funkhouser, T.; and Nie{\ss}ner,
  M. 2017.
\newblock Scannet: Richly-annotated 3d reconstructions of indoor scenes.
\newblock In \emph{CVPR}, 5828--5839.

\bibitem[{Fang, Easwaran, and Genest(2024)}]{fang2024uncertainty}
Fang, X.; Easwaran, A.; and Genest, B. 2024.
\newblock Uncertainty-guided appearance-motion association network for
  out-of-distribution action detection.
\newblock In \emph{MIPR}.

\bibitem[{Fang et~al.(2024{\natexlab{a}})Fang, Easwaran, Genest, and
  Suganthan}]{fang2024your}
Fang, X.; Easwaran, A.; Genest, B.; and Suganthan, P.~N. 2024{\natexlab{a}}.
\newblock Your Data Is Not Perfect: Towards Cross-Domain Out-of-Distribution
  Detection in Class-Imbalanced Data.
\newblock \emph{ESWA}.

\bibitem[{Fang et~al.(2025{\natexlab{a}})Fang, Easwaran, Genest, and
  Suganthan}]{fang2025adaptive}
Fang, X.; Easwaran, A.; Genest, B.; and Suganthan, P.~N. 2025{\natexlab{a}}.
\newblock Adaptive Hierarchical Graph Cut for Multi-granularity
  Out-of-distribution Detection.
\newblock \emph{IEEE TAI}.

\bibitem[{Fang et~al.(2024{\natexlab{b}})Fang, Fang, Liu, Qu, Dong, Zhou, Li,
  Xu, Chen, Zheng et~al.}]{fang2024not}
Fang, X.; Fang, W.; Liu, D.; Qu, X.; Dong, J.; Zhou, P.; Li, R.; Xu, Z.; Chen,
  L.; Zheng, P.; et~al. 2024{\natexlab{b}}.
\newblock Not all inputs are valid: Towards open-set video moment retrieval
  using language.
\newblock In \emph{ACM MM}.

\bibitem[{Fang et~al.(2024{\natexlab{c}})Fang, Fang, Wang, Liu, Tang, Dong,
  Zhou, and Li}]{fang2024multi}
Fang, X.; Fang, W.; Wang, C.; Liu, D.; Tang, K.; Dong, J.; Zhou, P.; and Li, B.
  2024{\natexlab{c}}.
\newblock Multi-Pair Temporal Sentence Grounding via Multi-Thread Knowledge
  Transfer Network.
\newblock \emph{arXiv}.

\bibitem[{Fang and Hu(2020)}]{fang2020double}
Fang, X.; and Hu, Y. 2020.
\newblock Double self-weighted multi-view clustering via adaptive view fusion.
\newblock \emph{arXiv}.

\bibitem[{Fang et~al.(2021{\natexlab{a}})Fang, Hu, Zhou, and
  Wu}]{fang2021animc}
Fang, X.; Hu, Y.; Zhou, P.; and Wu, D. 2021{\natexlab{a}}.
\newblock Animc: A soft approach for autoweighted noisy and incomplete
  multiview clustering.
\newblock \emph{IEEE TAI}.

\bibitem[{Fang et~al.(2020)Fang, Hu, Zhou, and Wu}]{fang2020v}
Fang, X.; Hu, Y.; Zhou, P.; and Wu, D.~O. 2020.
\newblock V3H: View variation and view heredity for incomplete multiview
  clustering.
\newblock \emph{IEEE TAI}.

\bibitem[{Fang et~al.(2021{\natexlab{b}})Fang, Hu, Zhou, and
  Wu}]{fang2021unbalanced}
Fang, X.; Hu, Y.; Zhou, P.; and Wu, D.~O. 2021{\natexlab{b}}.
\newblock Unbalanced incomplete multi-view clustering via the scheme of view
  evolution: Weak views are meat; strong views do eat.
\newblock \emph{IEEE TETCI}.

\bibitem[{Fang et~al.(2023{\natexlab{a}})Fang, Liu, Fang, Zhou, Cheng, Tang,
  and Zou}]{fang2023annotations}
Fang, X.; Liu, D.; Fang, W.; Zhou, P.; Cheng, Y.; Tang, K.; and Zou, K.
  2023{\natexlab{a}}.
\newblock Annotations Are Not All You Need: A Cross-modal Knowledge Transfer
  Network for Unsupervised Temporal Sentence Grounding.
\newblock In \emph{EMNLP}.

\bibitem[{Fang et~al.(2024{\natexlab{d}})Fang, Liu, Fang, Zhou, Xu, Xu, Chen,
  and Li}]{fang2024fewer}
Fang, X.; Liu, D.; Fang, W.; Zhou, P.; Xu, Z.; Xu, W.; Chen, J.; and Li, R.
  2024{\natexlab{d}}.
\newblock Fewer Steps, Better Performance: Efficient Cross-Modal Clip Trimming
  for Video Moment Retrieval Using Language.
\newblock In \emph{AAAI}.

\bibitem[{Fang et~al.(2022)Fang, Liu, Zhou, and Hu}]{fang2022multi}
Fang, X.; Liu, D.; Zhou, P.; and Hu, Y. 2022.
\newblock Multi-modal cross-domain alignment network for video moment
  retrieval.
\newblock \emph{IEEE TMM}.

\bibitem[{Fang et~al.(2023{\natexlab{b}})Fang, Liu, Zhou, and
  Nan}]{fang2023you}
Fang, X.; Liu, D.; Zhou, P.; and Nan, G. 2023{\natexlab{b}}.
\newblock You can ground earlier than see: An effective and efficient pipeline
  for temporal sentence grounding in compressed videos.
\newblock In \emph{CVPR}.

\bibitem[{Fang et~al.(2023{\natexlab{c}})Fang, Liu, Zhou, Xu, and
  Li}]{fang2023hierarchical}
Fang, X.; Liu, D.; Zhou, P.; Xu, Z.; and Li, R. 2023{\natexlab{c}}.
\newblock Hierarchical local-global transformer for temporal sentence
  grounding.
\newblock \emph{IEEE TMM}.

\bibitem[{Fang et~al.(2025{\natexlab{b}})Fang, Xiong, Fang, Qu, Chen, Dong,
  Tang, Zhou, Cheng, and Liu}]{fang2025rethinking}
Fang, X.; Xiong, Z.; Fang, W.; Qu, X.; Chen, C.; Dong, J.; Tang, K.; Zhou, P.;
  Cheng, Y.; and Liu, D. 2025{\natexlab{b}}.
\newblock Rethinking weakly-supervised video temporal grounding from a game
  perspective.
\newblock In \emph{ECCV}.

\bibitem[{Fei et~al.(2024{\natexlab{a}})Fei, Wu, Ji, Zhang, and
  Chua}]{fei2024dysen}
Fei, H.; Wu, S.; Ji, W.; Zhang, H.; and Chua, T.-S. 2024{\natexlab{a}}.
\newblock Dysen-VDM: Empowering Dynamics-aware Text-to-Video Diffusion with
  LLMs.
\newblock In \emph{CVPR}.

\bibitem[{Fei et~al.(2024{\natexlab{b}})Fei, Wu, Ji, Zhang, Zhang, Lee, and
  Hsu}]{fei2024video}
Fei, H.; Wu, S.; Ji, W.; Zhang, H.; Zhang, M.; Lee, M.-L.; and Hsu, W.
  2024{\natexlab{b}}.
\newblock Video-of-thought: Step-by-step video reasoning from perception to
  cognition.
\newblock In \emph{ICML}.

\bibitem[{Fei et~al.(2024{\natexlab{c}})Fei, Wu, Zhang, Chua, and
  Yan}]{fei2024vitron}
Fei, H.; Wu, S.; Zhang, H.; Chua, T.-S.; and Yan, S. 2024{\natexlab{c}}.
\newblock VITRON: A Unified Pixel-level Vision LLM for Understanding,
  Generating, Segmenting, Editing.
\newblock In \emph{NeurIPS}.

\bibitem[{Fei et~al.(2024{\natexlab{d}})Fei, Wu, Zhang, Zhang, Chua, and
  Yan}]{fei2024enhancing}
Fei, H.; Wu, S.; Zhang, M.; Zhang, M.; Chua, T.-S.; and Yan, S.
  2024{\natexlab{d}}.
\newblock Enhancing video-language representations with structural
  spatio-temporal alignment.
\newblock \emph{TPAMI}.

\bibitem[{Feng et~al.(2023)Feng, Liu, Liu, Zhou, Li, and
  Huang}]{feng2023docpedia}
Feng, H.; Liu, Q.; Liu, H.; Zhou, W.; Li, H.; and Huang, C. 2023.
\newblock Docpedia: Unleashing the power of large multimodal model in the
  frequency domain for versatile document understanding.
\newblock \emph{arXiv}.

\bibitem[{Garcia and Bruna(2017)}]{garcia2017few}
Garcia, V.; and Bruna, J. 2017.
\newblock Few-shot learning with graph neural networks.
\newblock \emph{arXiv preprint arXiv:1711.04043}.

\bibitem[{He and Ding(2024)}]{he2024refmask3d}
He, S.; and Ding, H. 2024.
\newblock RefMask3D: Language-guided transformer for 3D referring segmentation.
\newblock In \emph{ACM MM}, 8316--8325.

\bibitem[{He et~al.(2025)He, Ding, Jiang, and Wen}]{he2025segpoint}
He, S.; Ding, H.; Jiang, X.; and Wen, B. 2025.
\newblock Segpoint: Segment any point cloud via large language model.
\newblock In \emph{ECCV}, 349--367. Springer.

\bibitem[{He et~al.(2023)He, Jiang, Jiang, and Ding}]{he2023prototype}
He, S.; Jiang, X.; Jiang, W.; and Ding, H. 2023.
\newblock Prototype adaption and projection for few-and zero-shot 3d point
  cloud semantic segmentation.
\newblock \emph{IEEE TIP}, 32: 3199--3211.

\bibitem[{Hu et~al.(2024)Hu, Zhong, Liang, Zhang, Li, and Li}]{hu2024towards}
Hu, X.; Zhong, B.; Liang, Q.; Zhang, S.; Li, N.; and Li, X. 2024.
\newblock Towards Modalities Correlation for RGB-T Tracking.
\newblock \emph{IEEE TCSVT}.

\bibitem[{Hu et~al.(2023)Hu, Zhong, Liang, Zhang, Li, Li, and
  Ji}]{hu2023transformer}
Hu, X.; Zhong, B.; Liang, Q.; Zhang, S.; Li, N.; Li, X.; and Ji, R. 2023.
\newblock Transformer tracking via frequency fusion.
\newblock \emph{IEEE TCSVT}, 34(2): 1020--1031.

\bibitem[{Jiang et~al.(2023)Jiang, Wang, Ning, and Yu}]{jiang2023lttpoint}
Jiang, L.; Wang, C.; Ning, X.; and Yu, Z. 2023.
\newblock LTTPoint: A MLP-Based Point Cloud Classification Method with Local
  Topology Transformation Module.
\newblock In \emph{2023 7th Asian Conference on Artificial Intelligence
  Technology (ACAIT)}, 783--789. IEEE.

\bibitem[{Lei et~al.(2025)Lei, Cai, Liu, Fang, Qu, Dong, Yu, and
  Jin}]{lei2025exploring}
Lei, H.; Cai, X.; Liu, D.; Fang, X.; Qu, X.; Dong, J.; Yu, J.; and Jin, K.
  2025.
\newblock Exploring Disentangled Appearance-Motion Contexts for Temporal
  Activity Localization.
\newblock In \emph{IJCNN}.

\bibitem[{Liang et~al.(2024{\natexlab{a}})Liang, Feng, Zhou, Zhang, Zou, and
  Bai}]{liang2024pointgst}
Liang, D.; Feng, T.; Zhou, X.; Zhang, Y.; Zou, Z.; and Bai, X.
  2024{\natexlab{a}}.
\newblock Parameter-Efficient Fine-Tuning in Spectral Domain for Point Cloud
  Learning.
\newblock \emph{arXiv preprint arXiv:2410.08114}.

\bibitem[{Liang et~al.(2024{\natexlab{b}})Liang, Zhou, Xu, Zhu, Zou, Ye, Tan,
  and Bai}]{liang2024pointmamba}
Liang, D.; Zhou, X.; Xu, W.; Zhu, X.; Zou, Z.; Ye, X.; Tan, X.; and Bai, X.
  2024{\natexlab{b}}.
\newblock PointMamba: A Simple State Space Model for Point Cloud Analysis.
\newblock In \emph{NeurIPS}.

\bibitem[{Liu et~al.(2023{\natexlab{a}})Liu, Fang, Hu, and
  Zhou}]{liu2023exploring}
Liu, D.; Fang, X.; Hu, W.; and Zhou, P. 2023{\natexlab{a}}.
\newblock Exploring optical-flow-guided motion and detection-based appearance
  for temporal sentence grounding.
\newblock \emph{IEEE TMM}.

\bibitem[{Liu et~al.(2024{\natexlab{a}})Liu, Fang, Qu, Dong, Yan, Yang, Zhou,
  and Cheng}]{liu2024unsupervised}
Liu, D.; Fang, X.; Qu, X.; Dong, J.; Yan, H.; Yang, Y.; Zhou, P.; and Cheng, Y.
  2024{\natexlab{a}}.
\newblock Unsupervised domain adaptative temporal sentence localization with
  mutual information maximization.
\newblock In \emph{AAAI}.

\bibitem[{Liu et~al.(2023{\natexlab{b}})Liu, Fang, Zhou, Di, Lu, and
  Cheng}]{liu2023hypotheses}
Liu, D.; Fang, X.; Zhou, P.; Di, X.; Lu, W.; and Cheng, Y. 2023{\natexlab{b}}.
\newblock Hypotheses tree building for one-shot temporal sentence localization.
\newblock In \emph{AAAI}.

\bibitem[{Liu et~al.(2024{\natexlab{b}})Liu, Qu, Fang, Dong, Zhou, Nan, Tang,
  Fang, and Cheng}]{liu2024towards}
Liu, D.; Qu, X.; Fang, X.; Dong, J.; Zhou, P.; Nan, G.; Tang, K.; Fang, W.; and
  Cheng, Y. 2024{\natexlab{b}}.
\newblock Towards robust temporal activity localization learning with noisy
  labels.
\newblock In \emph{COLING}.

\bibitem[{Liu et~al.(2024{\natexlab{c}})Liu, Yang, Qu, Zhou, Fang, Tang, Wan,
  and Sun}]{liu2024pandora}
Liu, D.; Yang, M.; Qu, X.; Zhou, P.; Fang, X.; Tang, K.; Wan, Y.; and Sun, L.
  2024{\natexlab{c}}.
\newblock Pandora's Box: Towards Building Universal Attackers against
  Real-World Large Vision-Language Models.
\newblock In \emph{NeurIPS}.

\bibitem[{Liu et~al.(2023{\natexlab{c}})Liu, Zhu, Fang, Xiong, Wang, Li, and
  Zhou}]{liu2023conditional}
Liu, D.; Zhu, J.; Fang, X.; Xiong, Z.; Wang, H.; Li, R.; and Zhou, P.
  2023{\natexlab{c}}.
\newblock Conditional video diffusion network for fine-grained temporal
  sentence grounding.
\newblock \emph{IEEE TMM}.

\bibitem[{Liu et~al.(2024{\natexlab{d}})Liu, Yin, Wang, Chen, Sonke, and
  Gavves}]{liu2024dynamic}
Liu, J.; Yin, W.; Wang, H.; Chen, Y.; Sonke, J.-J.; and Gavves, E.
  2024{\natexlab{d}}.
\newblock Dynamic Prototype Adaptation with Distillation for Few-shot Point
  Cloud Segmentation.
\newblock In \emph{3DV}, 810--819. IEEE.

\bibitem[{Liu et~al.(2023{\natexlab{d}})Liu, Zhang, Peng, Huang, Wang, Tang,
  Huang, Lin, Shen, Bai et~al.}]{liu2023spts}
Liu, Y.; Zhang, J.; Peng, D.; Huang, M.; Wang, X.; Tang, J.; Huang, C.; Lin,
  D.; Shen, C.; Bai, X.; et~al. 2023{\natexlab{d}}.
\newblock Spts v2: single-point scene text spotting.
\newblock \emph{TPAMI}.

\bibitem[{Mao et~al.(2022)Mao, Guo, Xiaonan, Yuan, and
  Guo}]{mao2022bidirectional}
Mao, Y.; Guo, Z.; Xiaonan, L.; Yuan, Z.; and Guo, H. 2022.
\newblock Bidirectional feature globalization for few-shot semantic
  segmentation of 3d point cloud scenes.
\newblock In \emph{3DV}, 505--514. IEEE.

\bibitem[{Moody and Darken(1989)}]{moody1989fast}
Moody, J.; and Darken, C.~J. 1989.
\newblock Fast learning in networks of locally-tuned processing units.
\newblock \emph{Neural computation}, 1(2): 281--294.

\bibitem[{Ning et~al.(2023{\natexlab{a}})Ning, Wang, Zhang, Ning, and
  Tiwari}]{ning2023occluded}
Ning, E.; Wang, C.; Zhang, H.; Ning, X.; and Tiwari, P. 2023{\natexlab{a}}.
\newblock Occluded person re-identification with deep learning: a survey and
  perspectives.
\newblock \emph{ESWA}, 122419.

\bibitem[{Ning et~al.(2024)Ning, Wang, Wang, Zhang, and
  Ning}]{ning2024enhancement}
Ning, E.; Wang, Y.; Wang, C.; Zhang, H.; and Ning, X. 2024.
\newblock Enhancement, integration, expansion: Activating representation of
  detailed features for occluded person re-identification.
\newblock \emph{Neural Networks}, 169: 532--541.

\bibitem[{Ning et~al.(2023{\natexlab{b}})Ning, Zhang, Wang, Ning, Chen, and
  Bai}]{ning2023pedestrian}
Ning, E.; Zhang, C.; Wang, C.; Ning, X.; Chen, H.; and Bai, X.
  2023{\natexlab{b}}.
\newblock Pedestrian Re-ID based on feature consistency and contrast
  enhancement.
\newblock \emph{Displays}, 79: 102467.

\bibitem[{Ning et~al.(2023{\natexlab{c}})Ning, Tian, Lu, and
  Pei}]{ning2023boosting}
Ning, Z.; Tian, Z.; Lu, G.; and Pei, W. 2023{\natexlab{c}}.
\newblock Boosting few-shot 3d point cloud segmentation via query-guided
  enhancement.
\newblock In \emph{ACM MM}, 1895--1904.

\bibitem[{Park et~al.(2023)Park, Lee, Kim, Xiong, and Kim}]{park2023self}
Park, J.; Lee, S.; Kim, S.; Xiong, Y.; and Kim, H.~J. 2023.
\newblock Self-positioning Point-based Transformer for Point Cloud
  Understanding.
\newblock In \emph{CVPR}, 21814--21823.

\bibitem[{Qi et~al.(2017)Qi, Su, Mo, and Guibas}]{qi2017pointnet}
Qi, C.~R.; Su, H.; Mo, K.; and Guibas, L.~J. 2017.
\newblock Pointnet: Deep learning on point sets for 3d classification and
  segmentation.
\newblock In \emph{CVPR}, 652--660.

\bibitem[{Rosenblatt(1958)}]{rosenblatt1958perceptron}
Rosenblatt, F. 1958.
\newblock The perceptron: a probabilistic model for information storage and
  organization in the brain.
\newblock \emph{Psychological review}, 65(6): 386.

\bibitem[{Rudin et~al.(1964)}]{rudin1964principles}
Rudin, W.; et~al. 1964.
\newblock \emph{Principles of mathematical analysis}, volume~3.
\newblock McGraw-hill New York.

\bibitem[{Sereno et~al.(2020)Sereno, Wang, Besan{\c{c}}on, Mcguffin, and
  Isenberg}]{sereno2020collaborative}
Sereno, M.; Wang, X.; Besan{\c{c}}on, L.; Mcguffin, M.~J.; and Isenberg, T.
  2020.
\newblock Collaborative work in augmented reality: A survey.
\newblock \emph{IEEE TVCG}, 28(6): 2530--2549.

\bibitem[{Snell, Swersky, and Zemel(2017)}]{snell2017prototypical}
Snell, J.; Swersky, K.; and Zemel, R. 2017.
\newblock Prototypical networks for few-shot learning.
\newblock \emph{NeurIPS}, 30.

\bibitem[{Soori, Arezoo, and Dastres(2023)}]{soori2023artificial}
Soori, M.; Arezoo, B.; and Dastres, R. 2023.
\newblock Artificial intelligence, machine learning and deep learning in
  advanced robotics, a review.
\newblock \emph{Cognitive Robotics}, 3: 54--70.

\bibitem[{Tang et~al.(2023)Tang, Du, Wang, Zhou, Mei, Xue, Xu, and
  Zhang}]{tang2023character}
Tang, J.; Du, W.; Wang, B.; Zhou, W.; Mei, S.; Xue, T.; Xu, X.; and Zhang, H.
  2023.
\newblock Character recognition competition for street view shop signs.
\newblock \emph{National Science Review}.

\bibitem[{Tang et~al.(2024{\natexlab{a}})Tang, Lin, Zhao, Wei, Wu, Liu, Feng,
  Li, Wang, Liao et~al.}]{tang2024textsquare}
Tang, J.; Lin, C.; Zhao, Z.; Wei, S.; Wu, B.; Liu, Q.; Feng, H.; Li, Y.; Wang,
  S.; Liao, L.; et~al. 2024{\natexlab{a}}.
\newblock TextSquare: Scaling up Text-Centric Visual Instruction Tuning.
\newblock \emph{arXiv}.

\bibitem[{Tang et~al.(2022{\natexlab{a}})Tang, Qian, Song, Dong, Li, and
  Bai}]{tang2022optimal}
Tang, J.; Qian, W.; Song, L.; Dong, X.; Li, L.; and Bai, X. 2022{\natexlab{a}}.
\newblock Optimal boxes: boosting end-to-end scene text recognition by
  adjusting annotated bounding boxes via reinforcement learning.
\newblock In \emph{ECCV}.

\bibitem[{Tang et~al.(2022{\natexlab{b}})Tang, Qiao, Cui, Ma, Zhang, and
  Kanoulas}]{tang2022you}
Tang, J.; Qiao, S.; Cui, B.; Ma, Y.; Zhang, S.; and Kanoulas, D.
  2022{\natexlab{b}}.
\newblock You can even annotate text with voice: Transcription-only-supervised
  text spotting.
\newblock In \emph{ACM MM}.

\bibitem[{Tang et~al.(2022{\natexlab{c}})Tang, Zhang, Liu, Yang, Jiang, Hu, and
  Bai}]{tang2022few}
Tang, J.; Zhang, W.; Liu, H.; Yang, M.; Jiang, B.; Hu, G.; and Bai, X.
  2022{\natexlab{c}}.
\newblock Few could be better than all: Feature sampling and grouping for scene
  text detection.
\newblock In \emph{CVPR}.

\bibitem[{Tang et~al.(2025)Tang, Hou, Peng, Fang, Wu, Nie, Wang, and
  Tian}]{tang2025simplification}
Tang, K.; Hou, C.; Peng, W.; Fang, X.; Wu, Z.; Nie, Y.; Wang, W.; and Tian, Z.
  2025.
\newblock Simplification Is All You Need against Out-of-Distribution
  Overconfidence.
\newblock In \emph{CVPR}.

\bibitem[{Tang et~al.(2024{\natexlab{b}})Tang, Zhao, Peng, Fang, Cui, Zhu, and
  Tian}]{tang2024reparameterization}
Tang, K.; Zhao, W.; Peng, W.; Fang, X.; Cui, X.; Zhu, P.; and Tian, Z.
  2024{\natexlab{b}}.
\newblock Reparameterization head for efficient multi-input networks.
\newblock In \emph{ICASSP}.

\bibitem[{Wang, Cao, and Wang(2025)}]{wang2025learning}
Wang, C.; Cao, R.; and Wang, R. 2025.
\newblock Learning discriminative topological structure information
  representation for 2D shape and social network classification via persistent
  homology.
\newblock \emph{Knowledge-Based Systems}, 113125.

\bibitem[{Wang et~al.(2025{\natexlab{a}})Wang, Gao, Wu, Lam, He, and
  Tiwari}]{wang2025looking}
Wang, C.; Gao, X.; Wu, M.; Lam, S.-K.; He, S.; and Tiwari, P.
  2025{\natexlab{a}}.
\newblock Looking Clearer with Text: A Hierarchical Context Blending Network
  for Occluded Person Re-Identification.
\newblock \emph{TIFS}.

\bibitem[{Wang et~al.(2025{\natexlab{b}})Wang, He, Fang, Han, Liu, Ning, Li,
  and Tiwari}]{wang2025point}
Wang, C.; He, S.; Fang, X.; Han, J.; Liu, Z.; Ning, X.; Li, W.; and Tiwari, P.
  2025{\natexlab{b}}.
\newblock Point Clouds Meets Physics: Dynamic Acoustic Field Fitting Network
  for Point Cloud Understanding.
\newblock \emph{CVPR}.

\bibitem[{Wang et~al.(2023)Wang, Ning, Li, Bai, and Gao}]{wang20233d}
Wang, C.; Ning, X.; Li, W.; Bai, X.; and Gao, X. 2023.
\newblock 3D person re-identification based on global semantic guidance and
  local feature aggregation.
\newblock \emph{IEEE TCSVT}.

\bibitem[{Wang et~al.(2022{\natexlab{a}})Wang, Ning, Sun, Zhang, Li, and
  Bai}]{wang2022learning}
Wang, C.; Ning, X.; Sun, L.; Zhang, L.; Li, W.; and Bai, X. 2022{\natexlab{a}}.
\newblock Learning discriminative features by covering local geometric space
  for point cloud analysis.
\newblock \emph{IEEE TGRS}, 60: 1--15.

\bibitem[{Wang et~al.(2021)Wang, Wang, Li, and Wang}]{wang2021brief}
Wang, C.; Wang, C.; Li, W.; and Wang, H. 2021.
\newblock A brief survey on RGB-D semantic segmentation using deep learning.
\newblock \emph{Displays}, 70: 102080.

\bibitem[{Wang et~al.(2022{\natexlab{b}})Wang, Wang, Ning, Shengwei, and
  Li}]{wangchangshuo20223d}
Wang, C.; Wang, H.; Ning, X.; Shengwei, T.; and Li, W. 2022{\natexlab{b}}.
\newblock 3d point cloud classification method based on dynamic coverage of
  local area.
\newblock \emph{Journal of Software}, 34(4): 1962--1976.

\bibitem[{Wang et~al.(2024)Wang, Wu, Lam, Ning, Yu, Wang, Li, and
  Srikanthan}]{wang2024gpsformer}
Wang, C.; Wu, M.; Lam, S.-K.; Ning, X.; Yu, S.; Wang, R.; Li, W.; and
  Srikanthan, T. 2024.
\newblock GPSFormer: A Global Perception and Local Structure Fitting-based
  Transformer for Point Cloud Understanding.
\newblock \emph{arXiv preprint arXiv:2407.13519}.

\bibitem[{Wang et~al.(2025{\natexlab{c}})Wang, Lam, Wu, Hu, Wang, and
  Wang}]{wang2025destination}
Wang, R.; Lam, S.-K.; Wu, M.; Hu, Z.; Wang, C.; and Wang, J.
  2025{\natexlab{c}}.
\newblock Destination intention estimation-based convolutional encoder-decoder
  for pedestrian trajectory multimodality forecast.
\newblock \emph{Measurement}, 239: 115470.

\bibitem[{Wang et~al.(2019)Wang, Sun, Liu, Sarma, Bronstein, and
  Solomon}]{wang2019dynamic}
Wang, Y.; Sun, Y.; Liu, Z.; Sarma, S.~E.; Bronstein, M.~M.; and Solomon, J.~M.
  2019.
\newblock Dynamic graph cnn for learning on point clouds.
\newblock \emph{ACM TOG}, 38(5): 1--12.

\bibitem[{Xiong et~al.(2024)Xiong, Liu, Fang, Qu, Dong, Zhu, Tang, and
  Zhou}]{xiong2024rethinking}
Xiong, Z.; Liu, D.; Fang, X.; Qu, X.; Dong, J.; Zhu, J.; Tang, K.; and Zhou, P.
  2024.
\newblock Rethinking Video Sentence Grounding From a Tracking Perspective With
  Memory Network and Masked Attention.
\newblock \emph{IEEE TMM}.

\bibitem[{Yang et~al.(2025)Yang, Hou, Peng, Fang, Nie, Zhu, and
  Tang}]{yang2025eood}
Yang, G.; Hou, C.; Peng, W.; Fang, X.; Nie, Y.; Zhu, P.; and Tang, K. 2025.
\newblock EOOD: Entropy-based Out-of-distribution Detection.
\newblock In \emph{IJCNN}.

\bibitem[{Yu et~al.(2025)Yu, Wu, Lam, Wang, and Wang}]{yu2025eds}
Yu, S.; Wu, M.; Lam, S.-K.; Wang, C.; and Wang, R. 2025.
\newblock EDS-Depth: Enhancing Self-Supervised Monocular Depth Estimation in
  Dynamic Scenes.
\newblock \emph{TITS}.

\bibitem[{Yu et~al.(2024)Yu, Li, Xie, Wang, Li, and Ning}]{yu2024pedestrian}
Yu, Z.; Li, L.; Xie, J.; Wang, C.; Li, W.; and Ning, X. 2024.
\newblock Pedestrian 3d shape understanding for person re-identification via
  multi-view learning.
\newblock \emph{TCSVT}.

\bibitem[{Zhang et~al.(2023{\natexlab{a}})Zhang, Wu, Wu, Zhao, and
  Wang}]{zhang2023few}
Zhang, C.; Wu, Z.; Wu, X.; Zhao, Z.; and Wang, S. 2023{\natexlab{a}}.
\newblock Few-shot 3d point cloud semantic segmentation via stratified
  class-specific attention based transformer network.
\newblock In \emph{AAAI}, 3, 3410--3417.

\bibitem[{Zhang et~al.(2024{\natexlab{a}})Zhang, Ning, Wang, Ning, and
  Li}]{zhang2024deformation}
Zhang, H.; Ning, X.; Wang, C.; Ning, E.; and Li, L. 2024{\natexlab{a}}.
\newblock Deformation depth decoupling network for point cloud domain
  adaptation.
\newblock \emph{Neural Networks}, 180: 106626.

\bibitem[{Zhang et~al.(2023{\natexlab{b}})Zhang, Wang, Tian, Lu, Zhang, Ning,
  and Bai}]{zhang2023deep}
Zhang, H.; Wang, C.; Tian, S.; Lu, B.; Zhang, L.; Ning, X.; and Bai, X.
  2023{\natexlab{b}}.
\newblock Deep learning-based 3D point cloud classification: A systematic
  survey and outlook.
\newblock \emph{Displays}, 79: 102456.

\bibitem[{Zhang et~al.(2024{\natexlab{b}})Zhang, Wang, Yu, Tian, Ning, and
  Rodrigues}]{zhang2024pointgt}
Zhang, H.; Wang, C.; Yu, L.; Tian, S.; Ning, X.; and Rodrigues, J.
  2024{\natexlab{b}}.
\newblock PointGT: A Method for Point-Cloud Classification and Segmentation
  Based on Local Geometric Transformation.
\newblock \emph{IEEE TMM}.

\bibitem[{Zhang et~al.(2023{\natexlab{c}})Zhang, Wang, Wang, Gao, Li, and
  Shi}]{zhang2023starting}
Zhang, R.; Wang, L.; Wang, Y.; Gao, P.; Li, H.; and Shi, J. 2023{\natexlab{c}}.
\newblock Starting from non-parametric networks for 3d point cloud analysis.
\newblock In \emph{CVPR}, 5344--5353.

\bibitem[{Zhang et~al.(2025{\natexlab{a}})Zhang, Lei, Liu, Qu, Fang, Guan, and
  Jin}]{zhang2025manipulating}
Zhang, X.; Lei, H.; Liu, D.; Qu, X.; Fang, X.; Guan, R.; and Jin, K.
  2025{\natexlab{a}}.
\newblock Manipulating the Bounding Box: Multimodal Controlled Backdoor Attacks
  on 3D Visual Grounding Models.
\newblock In \emph{IJCNN}.

\bibitem[{Zhang et~al.(2025{\natexlab{b}})Zhang, Lei, Liu, Qu, Fang, Guan, and
  Jin}]{zhang2025monoAttack}
Zhang, X.; Lei, H.; Liu, D.; Qu, X.; Fang, X.; Guan, R.; and Jin, K.
  2025{\natexlab{b}}.
\newblock MonoAttack: A Strong Attack Framework with Depth-Migration and
  Attribute-Tampering for Monocular 3D Object Detection.
\newblock In \emph{IJCNN}.

\bibitem[{Zhao, Chua, and Lee(2021)}]{zhao2021few}
Zhao, N.; Chua, T.-S.; and Lee, G.~H. 2021.
\newblock Few-shot 3d point cloud semantic segmentation.
\newblock In \emph{CVPR}, 8873--8882.

\bibitem[{Zhao et~al.(2024{\natexlab{a}})Zhao, Tang, Lin, Wu, Huang, Liu, Tan,
  Zhang, and Xie}]{zhao2024multi}
Zhao, Z.; Tang, J.; Lin, C.; Wu, B.; Huang, C.; Liu, H.; Tan, X.; Zhang, Z.;
  and Xie, Y. 2024{\natexlab{a}}.
\newblock Multi-modal In-Context Learning Makes an Ego-evolving Scene Text
  Recognizer.
\newblock In \emph{CVPR}.

\bibitem[{Zhao et~al.(2024{\natexlab{b}})Zhao, Tang, Wu, Lin, Wei, Liu, Tan,
  Zhang, Huang, and Xie}]{zhao2024harmonizing}
Zhao, Z.; Tang, J.; Wu, B.; Lin, C.; Wei, S.; Liu, H.; Tan, X.; Zhang, Z.;
  Huang, C.; and Xie, Y. 2024{\natexlab{b}}.
\newblock Harmonizing Visual Text Comprehension and Generation.
\newblock \emph{arXiv}.

\bibitem[{Zhou et~al.(2024)Zhou, Liang, Xu, Zhu, Xu, Zou, and
  Bai}]{zhou2024dynamic}
Zhou, X.; Liang, D.; Xu, W.; Zhu, X.; Xu, Y.; Zou, Z.; and Bai, X. 2024.
\newblock Dynamic Adapter Meets Prompt Tuning: Parameter-Efficient Transfer
  Learning for Point Cloud Analysis.
\newblock In \emph{CVPR}, 14707--14717.

\bibitem[{Zhu et~al.(2023)Zhu, Zhou, Yao, and Zhu}]{zhu2023cross}
Zhu, G.; Zhou, Y.; Yao, R.; and Zhu, H. 2023.
\newblock Cross-class bias rectification for point cloud few-shot segmentation.
\newblock \emph{IEEE TMM}, 25: 9175--9188.

\bibitem[{Zhu et~al.(2024)Zhu, Zhang, He, Guo, Liu, Xiao, Fu, Dong, and
  Gao}]{zhu2024no}
Zhu, X.; Zhang, R.; He, B.; Guo, Z.; Liu, J.; Xiao, H.; Fu, C.; Dong, H.; and
  Gao, P. 2024.
\newblock No Time to Train: Empowering Non-Parametric Networks for Few-shot 3D
  Scene Segmentation.
\newblock In \emph{CVPR}, 3838--3847.

\end{thebibliography}

\end{document}